\documentclass[sigconf, screen]{acmart} 

\AtBeginDocument{%
  }

\setcopyright{acmlicensed}
\copyrightyear{2025}
\acmYear{2025}
\acmDOI{XXXXXXX.XXXXXXX}

\acmConference[Conference acronym 'XX]{Make sure to enter the correct
  conference title from your rights confirmation email}{June XX--XX,
  2025}{Woodstock, NY}

\acmISBN{XXX-X-XXXX-XXXX-X/XXXX/XX}
\usepackage{flushend}
\usepackage{balance}
\usepackage{booktabs}
\usepackage{multirow} 
\usepackage{subfigure}
\usepackage{diagbox}
  
\usepackage{amsmath,amsthm,amssymb,amsfonts}
\usepackage{algorithm}
\usepackage{algorithmic}
\usepackage{pifont}
\usepackage{bbding}
\usepackage[most]{tcolorbox}

\begin{document}

\title{RAMA: Retrieval-Augmented Multi-Agent Framework for Misinformation Detection in Multimodal Fact-Checking}

\author{Shuo Yang}
\affiliation{
  \institution{The University of Hong Kong}
  \city{Hong Kong SAR}
  \country{China}
}

\author{Zijian Yu}
\affiliation{
  \institution{Ant Group}
  \city{Hangzhou}
  \country{China}}

\author{Zhenzhe Ying}
\affiliation{
  \institution{Ant Group}
  \city{Hangzhou}
  \country{China}
}

\author{Yuqin Dai}
\affiliation{
 \institution{Ant Group}
 \city{Hangzhou}
 \country{China}}

\author{Guoqing Wang}
\affiliation{
  \institution{Ant Group}
  \city{Hangzhou}
  \country{China}}

\author{Jun Lan}
\affiliation{
  \institution{Ant Group}
  \city{Hangzhou}
  \country{China}}

\author{Jinfeng Xu}
\affiliation{%
  \institution{The University of Hong Kong}
  \city{Hong Kong SAR}
  \country{China}
}

\author{Jinze Li}
\affiliation{
  \institution{The University of Hong Kong}
  \city{Hong Kong SAR}
  \country{China}
}

\author{Edith C.H. Ngai}
\affiliation{
  \institution{The University of Hong Kong}
  \city{Hong Kong SAR}
  \country{China}}

\renewcommand{\shortauthors}{Yang et al.}

\begin{abstract}

The rapid proliferation of multimodal misinformation presents significant challenges for automated fact-checking systems, especially when claims are ambiguous or lack sufficient context. We introduce RAMA, a novel retrieval-augmented multi-agent framework designed for verifying multimedia misinformation. RAMA incorporates three core innovations: (1) strategic query formulation that transforms multimodal claims into precise web search queries; (2) cross-verification evidence aggregation from diverse, authoritative sources; and (3) a multi-agent ensemble architecture that leverages the complementary strengths of multiple multimodal large language models and prompt variants. Extensive experiments demonstrate that RAMA achieves superior performance on benchmark datasets, particularly excelling in resolving ambiguous or improbable claims by grounding verification in retrieved factual evidence. Our findings underscore the necessity of integrating web-based evidence and multi-agent reasoning for trustworthy multimedia verification, paving the way for more reliable and scalable fact-checking solutions. RAMA will be publicly available at \href{}{https://github.com/kalendsyang/RAMA.git}.
\end{abstract}



\maketitle

\section{Introduction}

In the digital era, the explosive growth of information has greatly facilitated knowledge acquisition, but at the same time, it has also accelerated the spread of false and misleading content, particularly in critical domains such as healthcare, politics, and economics~\cite{augenstein2024factuality}. Misinformation risks severe societal consequences, including harmful medical practices~\cite{okati2024truth}, economic volatility~\cite{meel2020fake}, and the erosion of public trust~\cite{huan2024social}. As the primary conduit for information distribution, social media platforms exacerbate this crisis by lowering content-creation barriers and amplifying propagation speed~\cite{islam2020deep, zhang2024toward}. Therefore, efficiently verifying and ensuring the accuracy of information has become an urgent task in combating misinformation.

Among prevalent deceptive media, deepfakes (AI-synthesized multimedia content)~\cite{masood2023deepfakes, jia2024can} and cheapfakes (low-cost manipulations using tools like Photoshop for editing or out-of-context misuse)~\cite{paris2019deepfakes, vo2024detecting, vu2024enhancing} constitute two major threats. Compared to deepfakes, cheapfakes are particularly hazardous due to their ease of creation, wide dissemination, and high degree of concealment. The representative technique of cheapfakes, known as out-of-context (OOC) misuse, involves pairing unaltered images with misleading captions~\cite{papadopoulos2025similarity, alkaddour2022sentiment}. Since the images remain unmodified, traditional detection methods based on visual tampering traces are ineffective, making the verification of OOC image-caption pairs a significant research challenge. In recent years, the detection of OOC multimedia content has emerged as a key research focus, spurring the development of dedicated competitions~\cite{dang2023grand, nguyen2024overview, MV2025overview} and datasets~\cite{aneja2021cosmos, luo2021newsclippings, xu2024mmooc}.

Previous detection methods predominantly employ deep learning paradigms, including supervised learning~\cite{alkaddour2022sentiment}, self-supervised learning~\cite{tran2022textual}, generative data augmentation~\cite{nguyen2024unified, vo2024detecting}, and pretrained image-text models~\cite{le2024tega}. However, these approaches suffer from two critical limitations: (1) Poor generalization: Models tend to overfit limited training data and struggle to adapt to real-world scenarios~\cite{nguyen2024unified}; (2) Lack of interpretability: Black-box decision mechanisms fail to provide users with credible reasoning, which hinders practical application~\cite{qi2024sniffer}. Recent advances in large language models (LLMs) offer new opportunities for automated misinformation detection. By leveraging extensive knowledge bases, LLMs demonstrate capabilities in contextual reasoning~\cite{zhou2025towards, wang2025decoupling}, evidence retrieval~\cite{ma2025local, tang2024minicheck}, and explainable decision generation~\cite{yang2025largelanguagemodelsnetwork, warren2025show}, which have been validated in prior efforts~\cite{wu2023cheap, vu2024enhancing}. Nevertheless, LLMs still encounter challenges in OOC scenarios, particularly regarding knowledge recency and the capture of dynamic contexts with cross-modal semantic consistency.

To address these issues, this paper proposes a Retrieval-Augmented Multi-Agent (RAMA) framework designed for multimodal misinformation detection. Inspired by human fact-checking workflows~\cite{warren2025show}, RAMA integrates retrieval-augmented verification with multi-agent collaboration. Its core components include: the WebRetriever module, which dynamically retrieves and collects evidence in realistic web environment to provide contextual information for decision-making; the VLJudge module, composed of three independent multimodal large language model (MLLM) agents, responsible for evaluating the consistency between the input image, caption, and the retrieved context; and the DecisionFuser, which aggregates the judgments of agents through a majority voting mechanism, thereby enhancing the robustness and scalable of the system. Through multi-agent collaboration and evidence-driven reasoning, RAMA effectively overcomes the limitations of existing methods in terms of data dependency, generalization ability, and interpretability, providing a new paradigm for multimedia misinformation detection.

In summary, our contributions are: 
\begin{itemize}
    \item We propose RAMA, a novel retrieval-augmented multi-agent framework for misinformation detection. RAMA integrates web-based evidence retrieval with multi-agent reasoning, outperforming existing approaches.

    \item We introduce a WebRetriever module that decomposes complex multimodal claims and transforms them into precise, context-aware web search queries. This enables the effective retrieval and cross-verification of relevant and authoritative evidence.

    \item We develop a multi-agent ensemble verification architecture that leverages the complementary strengths of multiple MLLMs and diverse prompt strategies. This design mitigates individual agent noise and bias, substantially enhancing overall verification reliability.

    \item We conduct extensive quantitative and qualitative experiments, demonstrating that RAMA achieves state-of-the-art performance and robustly resolves challenging real-world misinformation scenarios.
\end{itemize}

\section{Related Work}

\subsection{Misinformation Detection}

Social media platforms like Facebook, Twitter, and Weibo have become deeply integrated into daily life, serving as vital channels for users to share personal information, images, and videos. However, with their widespread adoption, the internet is flooded with a significant amount of inaccurate, misleading, and even false information. Such deceptive content, including fake news and rumors, not only misleads users but also constitutes a global risk~\cite{islam2020deep}. To address this challenge, misinformation detection technologies~\cite{yu2017convolutional, zhang2016misinformation, khattar2019mvae} have emerged, leveraging artificial intelligence (AI) and machine learning (ML) to automatically identify, classify, and respond to harmful information.

Traditional methods for detecting misinformation primarily focus on content analysis~\cite{bondielli2019survey}, typically deconstructing original content into different granularities~\cite{pelrine2021surprising}. At the word level, they analyze statistical features such as word frequency and word embeddings~\cite{huang2020conquering, qian2018neural, botnevik2020brenda}. At the sentence level, they consider syntactic features and complexity~\cite{shu2019defend, cheng2020vroc, ito2015assessment}. On a broader scale, they examine text representation, topic distribution, and sentiment tendencies~\cite{kumari2021misinformation, apostol2024contcommrtd, wan2024dell}. In the era of multimedia information, a prevalent deceptive strategy involves placing unaltered real images in misleading textual contexts, known as out-of-context (OOC) misinformation~\cite{qi2024sniffer, papadopoulos2025similarity, abdali2024multi}. Also referred to as cheapfakes~\cite{paris2019deepfakes}, these are created using simple editing tools or misleading captions, in contrast to deepfakes, which utilize advanced generative models. The OOC issue is inherently multimodal, requiring the integration of natural language processing and computer vision technologies for detection. Some approaches~\cite{zhou2023multimodal,mu2023self} leverage pretrained models, such as CLIP~\cite{radford2021learningtransferablevisualmodels}, to directly assess the internal consistency of image-text pairs, while others~\cite{shang2024mmadapt, qian2021knowledge, jaiswal2019aird} utilize external resources for cross-validation. For example, researchers~\cite{shang2024mmadapt} have employed a local knowledge base containing unaltered relevant statements as a trusted source, comparing target statements with retrieved external information to identify health misinformation. Recent Studies~\cite{abdelnabi2022open, papadopoulos2025red} indicate that incorporating external information significantly enhances detection performance.

OOC misinformation detection faces unique challenges because the visual content is authentic and credible, while the deception stems from its association with erroneous text. This makes traditional content analysis methods less effective, consequently attracting widespread attention and driving the development of relevant datasets~\cite{luo2021newsclippings, aneja2021cosmos, xu2024mmooc}, algorithms~\cite{wu2023cheap, nguyen2024unified, seo2024multi}, and benchmark competitions~\cite{dang2023grand, nguyen2024overview, MV2025overview}. Recent advances include methods that utilize synthetic data and generative models. Nguyen~\textit{et al.}~\cite{nguyen2024unified} proposed a unified end-to-end network that leverages synthetic data generation for model training. Le \textit{et al.}~\cite{le2024tega} introduced a diffusion-based approach to merge the content of original images and captions, creating context-synthesized images. The differences between these and the original images are then quantified using a CILP model to distinguish OOC from non-OOC samples. Seo~\textit{et al.}~\cite{seo2024multi} enhanced both textual and visual modalities using a BERT-based transformer and stable diffusion, employing similarity comparison for classification. With the rapid advancement of large language models (LLMs), recent works have explored their potential for cheapfake detection. Wu~\textit{et al.}~\cite{wu2023cheap} developed a robust feature extractor based on GPT-3.5, capturing the correlation between multiple captions and integrating this module into a classification framework. Vo-Hoang \textit{et al.}~\cite{vo2024detecting} and Pham~\textit{et al.}~\cite{pham2024generative} utilized LLaMA-Adapter V2 for data augmentation and either fine-tuned or prompted parameter-efficient models for caption classification. Vu~\textit{et al.}~\cite{vu2024enhancing} designed a set of questions for LLaMA, aggregating integer outputs to determine OOC risks via thresholding. 

Despite significant progress with these methods, current research still exhibits notable limitations. On the one hand, utilizing generative models for data augmentation may cause models to overfit to the distribution of synthetic data. On the other hand, the lack of access to real-time dynamic information and an over-reliance on static knowledge bases or the parameterized knowledge embedded in LLMs restricts the model's ability to effectively reflect and adapt to the rapidly evolving contexts of the real world, thereby limiting its detection efficacy in dynamic scenarios.

\subsection{Multimodal Fact-Checking}

Fact-checking aims to determine the veracity or potential manipulation of a given claim~\cite{guo2022survey}, which is crucial in domains such as cheapfake detection, rumor verification, and fake news identification. While early fact-checking systems primarily relied on textual analysis~\cite{Wadden2020, Saakyan2021,tang2024minicheck, choi2024automated}, recent advances have increasingly incorporated multimodal information—including images~\cite{yao2023end, wang2024mfc, yang2025realfactbench, papadopoulos2025red} and videos~\cite{rayar2022large, micallef2022cross,cao2025video}—to evaluate cross-modal consistency and to integrate heterogeneous evidence for more robust justification. Interpretability has emerged as a critical requirement, as transparent and explainable predictions foster greater human trust and understanding of system outputs~\cite{yao2023end,warren2025show}. LLMs have demonstrated considerable potential for automating both detection and explanation in fact-checking. However, existing LLM-based systems~\cite{choi2024fact, li2024large, ma2025local} are often constrained by reliance on outdated pretraining corpora and limited exploitation of multimodal cues. Drawing inspiration from the workflow of human fact-checkers~\cite{warren2025show}, which encompasses evidence retrieval, cross-verification, expert collaboration, and explainable decision-making, we propose a multimodal fact-checking framework grounded in LLMs and MLLMs. Our approach enables automated reasoning, strategic planning, and the dynamic invocation of web retrieval tools to acquire up-to-date factual evidence, thereby facilitating rigorous cross-modal verification of image-text consistency. Experiments on a benchmark OOC misinformation detection dataset demonstrate the effectiveness of our proposed method.

\section{Methodology}

\subsection{Overview of RAMA}

\begin{figure}
    \centering
    \includegraphics[width=1\linewidth]{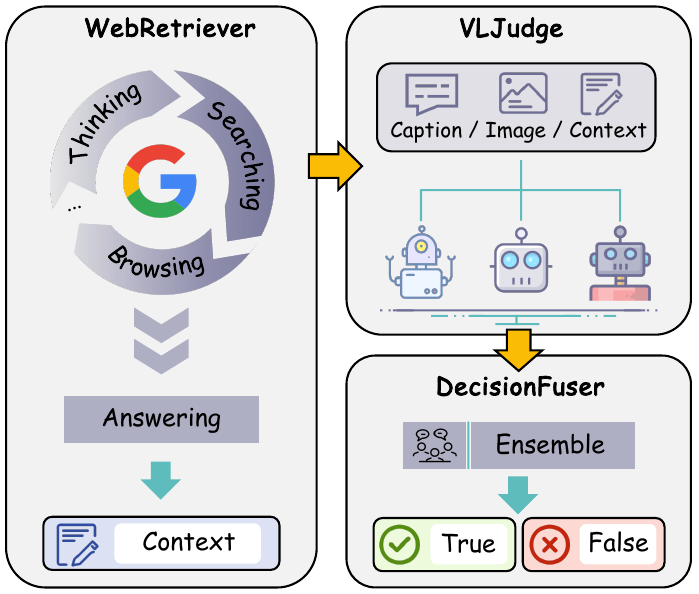}
    \caption{Overview of the RAMA workflow. RAMA consists of three cascaded modules: WebRetriever, VLJudge, and DecisionFuser.}
    \label{fig: overview}
\end{figure}

RAMA (Retrieval-Augmented Multi-Agent Framework) is a novel multi-agent system designed for misinformation detection in multimedia content. Inspired by human fact-checking workflows, RAMA seamlessly integrates retrieval-augmented verification with collaborative multi-agent reasoning to assess the authenticity of images paired with textual claims. As illustrated in Figure~\ref{fig: overview}, the framework comprises three core modules that operate in a cascaded manner to ensure comprehensive verification. WebRetriever first receives multimodal inputs and performs evidence retrieval using web-based tools to gather up-to-date external information. The retrieved content is then processed by VLJudge, which performs cross-modal verification and multi-agent reasoning to evaluate image-text consistency and generate interpretable decisions. Finally, DecisionFuser aggregates the judgments from all agents, further enhancing the robustness and scalability of the system. 

\subsection{WebRetriever: Contextual Evidence Gathering}
The WebRetriever module serves as the foundational evidence collection component, addressing the critical need for real-world contextual grounding in fake detection. Unlike conventional approaches that rely solely on pretrained knowledge, our system actively retrieves relevant descriptions from open-domain knowledge bases (including news archives, social media platforms, and encyclopedic sources) to establish an information baseline for subsequent verification.

The retrieval process operates in an iterative, tool-augmented reasoning loop, consisting of the following stages:

\paragraph{Thinking}: Before taking action, WebRetriever performs explicit reasoning, encapsulated in a <think> block. This step enables the agent to decompose complex queries, plan evidence collection strategies, and improve retrieval relevance.

\paragraph{Searching}: The agent generates a JSON-formatted request to invoke a web search tool, specifying the search query. The search tool returns the top-k (k=10) results, each including the title, URL, and summary of the webpage.

\paragraph{Browsing}: For each retrieved URL, WebRetriever issues a browsing request to extract novel content not previously encountered in the dialogue history. A short-term memory buffer is maintained for each query to avoid redundant information and enhance evidence diversity.

\paragraph{Answering}: Once sufficient information is gathered, the agent aggregates and synthesizes the evidence into a concise, objective summary, encapsulated in an <answer> block. This summary serves as the unified context for downstream consistency evaluation, reducing noise and redundancy.

Formally, let $\mathcal{T}$ denote the input textual claim (caption), $\mathcal{E} = \{e_1, e_2, ..., e_n\}$ denote the set of retrieved evidence. The context summary $\mathcal{C}$ is generated by a summarization function $f_{sum}$: 
\begin{equation}
    \mathcal{C} = f_{sum}(\mathcal{E}).
\end{equation}

\subsection{VLJudge: Multi-Agent Consistency Assessment}

The VLJudge module consists of three independent MLLM agents, denoting $\mathbb{J}=\{\mathcal{J}_1, \mathcal{J}_2, \mathcal{J}_3\}$, each tasked with evaluating the consistency between vision and language modality. This ensemble approach leverages model diversity to mitigate individual model bias and enhance robustness.

Each MLLM agent receives as input the base64-encoded image, the caption, and the context summary. The agents are prompted to assess three primary categories of cheapfake manipulation: (1) Image Manipulation: Detecting direct visual alterations (e.g., color changes, object insertion/removal). (2) Text Distortion: Identifying misleading or inaccurate textual claims that misrepresent the factual information. (3) Image-Text Mismatch: Identifying semantic inconsistencies between the image content and the textual claim.

For each agent $\mathcal{J}_i \in \mathbb{J}$, the output is categorized verdict $v_i \in \{\text{True}, \text{False}\}$, where:
\begin{itemize}
    \item True: The image and caption are consistent with the retrieved context.
    \item False: The image is manipulated or the caption is inconsistent with the image/context.
\end{itemize}

We design agent-specific prompts to elicit trustworthy and interpretable outputs. We considered the currently outstanding open-source MLLM as the judge, with $\mathcal{J}_1$ and $\mathcal{J}_3$ using Qwen-VL, and $\mathcal{J}_2$ using InternVL. For each image-caption pair ($\{\mathcal{T}, \mathcal{I}\}$) and its corresponding evidence $\mathbf{C}$, the agent’s decision is: 

\begin{equation}
    v_i = \mathcal{J}_i(\{\mathcal{T}, \mathcal{I}\}|\mathcal{C}), \quad i=1,2,3.
\end{equation}

Notably, we allow $\mathcal{J}_3$ to output "unknown," which avoids the model's random guesses for borderline cases. By applying different prompts and different MLLMs, we can encourage the model to focus on distinct aspects of the verification task, further enriching the ensemble's decision space. 

\subsection{DecisionFuser: Ensemble-Based Decision Making}

The DecisionFuser aggregates independent verdicts from the VLJudge agents and produces the final detection result using a majority voting strategy. This ensemble mechanism leverages the strengths of multiple MLLMs, mitigating the influence of individual model noise and bias, thereby enhancing overall reliability.

To further increase flexibility, the final module employs a weighted voting mechanism that synthesizes the judges' outputs into a consolidated decision. Let $\mathcal{W} = [w_1, w_2, w_3]$ represent the reliability weights for each judge (empirically determined from validation performance). The final decision is computed as:
\begin{equation}
    v = \operatorname{argmax}_{l \in \{\text{True}, \text{False}\}} \sum_{i=1}^3 w_i \cdot \mathbb{I}(v_i = l),
\end{equation}
where $\mathbb{I}(\cdot)$ is the indicator function. For simplicity, we map "unknown" to False. In practical scenarios, RAMA can seamlessly integrate with existing fact-checking systems. When the voting results are inconclusive or "unknown" is the majority, the uncertain case could be marked for further review.

\section{Experiment}

\subsection{Dataset and Evaluation Metrics}

For empirical evaluation, we utilized the official datasets provided by the Multimedia Verification Challenge\footnote{\url{https://multimedia-verification.github.io/}}. As our approach is non-parametric and does not require model training, all experiments were conducted exclusively on the public test set. This test set comprises 1,000 samples, each consisting of an image, two candidate captions, and a ground-truth label indicating either OOC (Out-of-Context) or NOOC (Not Out-of-Context). Following the competition guidelines, only the image and the first caption were used for prediction.

We adopted five standard evaluation metrics as specified by the challenge: Accuracy (Acc), Precision (P), Recall (R), F1-score (F1), and Matthews correlation coefficient (MCC). The definitions of these metrics are as follows:

\begin{align} &\text{Accuracy} = \frac{\text{TP} + \text{TN}}{\text{TP} + \text{TN} + \text{FP} + \text{FN}}, \\ &\text{Precision} = \frac{\text{TP}}{\text{TP} + \text{FP}}, \\ &\text{Recall} = \frac{\text{TP}}{\text{TP} + \text{FN}}, \\ &\text{F1-score} = \frac{2 \times \text{Precision} \times \text{Recall}}{\text{Precision} + \text{Recall}}, \\ &\text{MCC} = \frac{\text{TP} \times \text{TN} - \text{FP} \times \text{FN}}{\sqrt{(\text{TP} + \text{FP})(\text{TP} + \text{FN})(\text{TN} + \text{FP})(\text{TN} + \text{FN})}}. \end{align}

Here, $\text{TP}$ denotes the number of image-caption pairs correctly classified as OOC, $\text{TN}$ denotes those correctly classified as NOOC, $\text{FP}$ represents pairs incorrectly classified as OOC, and $\text{FN}$ denotes pairs incorrectly classified as NOOC. Precision, Recall, and F1-score are computed with respect to the "False" (OOC) class, aligning with the evaluation protocol of the competition.

\subsection{Implementation Details}
\label{sec: imp}

All experiments were conducted on a NVIDIA Tesla A100 80GB GPU (CUDA 12.4) using PyTorch (v2.5.1). Our pipeline integrates several state-of-the-art open-source models, all sourced from the Hugging Face repository. Specifically, the WebRetriever module employs the GAIR/DeepResearcher-7b model~\cite{zheng2025deepresearcher} for evidence retrieval. For cross-modal verification, the VLJudge module utilizes AWQ-quantized versions of Qwen2.5-VL-72B~\cite{bai2025qwen2} and InternVL3-78B~\cite{zhu2025internvl3}, significantly reducing memory consumption while maintaining performance. To accelerate inference, we adopt vLLM (v0.7.3) for efficient large language model serving. All evaluations are performed in a strict zero-shot setting, without any model fine-tuning or parameter updates. Task adaptation is achieved solely through prompt engineering, ensuring fair and reproducible assessment of model generalization.

\subsection{Comparison Results}

We report the performance of RAMA on the public test set and compare it with state-of-the-art methods using accuracy and F1-score as evaluation metrics. The results of comparison methods are borrowed from~\cite{nguyen2024overview}.

\begin{table}[t]
    \centering
    \caption{Comparison results on the public test set. Our results are highlighted in bold. Methods marked with * utilize both caption1 and caption2.}
    \label{tab:cmp_with_other}
    \begin{tabular}{ccc}
         \toprule
         \textbf{Method} & \textbf{Accuracy} & \textbf{F1-score} \\
         \midrule
         Pham \textit{et al.}*~\cite{pham2024generative} & 0.8890 & 0.8890 \\
         Vo-Hoang \textit{et al.}*~\cite{vo2024detecting} & 0.8600 & 0.8600 \\
         Le \textit{et al.}*~\cite{le2024tega} & 0.7940 & 0.7240 \\
         Vu \textit{et al.}*~\cite{vu2024enhancing} & 0.8290 & 0.8213 \\
         Nguyen \textit{et al.}~\cite{nguyen2024unified} & 0.9300 & 0.9260 \\
         Seo \textit{et al.}~\cite{seo2024multi} & 0.5570 & 0.6370 \\
         \textbf{RAMA(ours)} & \textbf{0.9100} & \textbf{0.9100} \\
         \bottomrule
    \end{tabular}
\end{table}

As summarized in Table~\ref{tab:cmp_with_other}, several prior approaches~\cite{pham2024generative, vo2024detecting, le2024tega, vu2024enhancing} utilize both provided captions (caption1 and caption2) for decision-making, whereas Nguyen \textit{et al.}~\cite{nguyen2024unified}, Seo \textit{et al.}~\cite{seo2024multi}, and our RAMA framework rely solely on caption1, in accordance with the competition guidelines. Nguyen \textit{et al.}~\cite{nguyen2024unified} achieved the highest accuracy (0.930) and F1-score (0.926) on the public test set. However, their method exhibited a substantial performance drop on the 2024 private test set (accuracy: 0.655, F1-score: 0.5174)~\cite{nguyen2024overview}, indicating potential overfitting and limited generalization.

Compared to the other methods, our RAMA approach achieves competitive results without any training or fine-tuning, attaining an accuracy of 0.910 and an F1-score of 0.910. By leveraging retrieved external evidence to support multimedia verification, RAMA demonstrates reduced sensitivity to dataset distribution and exhibits stronger generalization across unseen samples. 

\begin{figure*}
    \centering
    \includegraphics[width=1\linewidth]{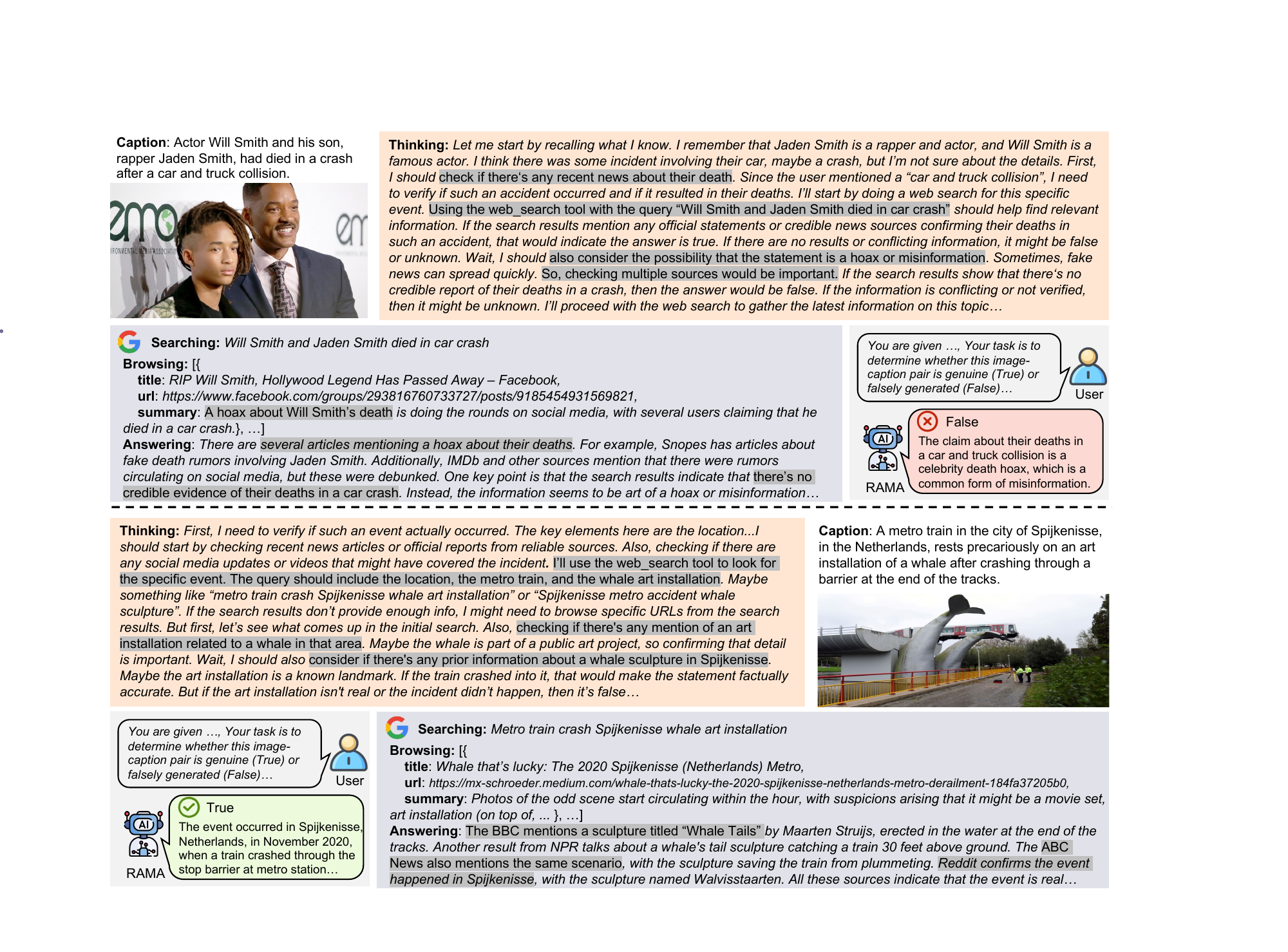}
    \caption{Case Study. RAMA’s verification process on challenging multimodal misinformation claims.}
    \label{fig: case}
\end{figure*}

\subsection{Ablation Study}

To systematically assess the contribution of key components in our framework, we conducted ablation experiments focusing on the impact of external evidence and model ensemble strategies. The quantitative results are summarized in Table~\ref{tab:ablation_study}.

\begin{table}[t]
    \setlength{\tabcolsep}{4pt}
    \centering
    \caption{Ablation study of prompt engineering and evidence integration on verification performance. P1/P2: different prompt templates; E: evidence retrieved by WebRetriever.}
    \label{tab:ablation_study}
    \begin{tabular}{cccccc}
         \toprule
         \textbf{Method} & \textbf{Acc} & \textbf{P} & \textbf{R} & \textbf{F1} & \textbf{MCC} \\
         \midrule
         QwenVL P1 w/o E & 0.7608 & 0.8503 & 0.6360 & 0.7277 & 0.5398 \\
         QwenVL P2 w/o E & 0.8330 & 0.8124 & 0.8660 & 0.8383 & 0.6675 \\
         InternVL P2 w/o E & 0.8670 & 0.9105 & 0.8140 & 0.8596 & 0.7382 \\
         \midrule
         QwenVL P1 w/ E & 0.8150 & 0.7818 & 0.8740 & 0.8253 & 0.6344 \\
         QwenVL P2 w/ E & 0.8910 & \textbf{0.9453} & 0.8300 & 0.8839 & 0.7879 \\
         InternVL P2 w/ E & 0.9000 & 0.9167 & 0.8800 & 0.8907 & 0.7842 \\
         \midrule
         \textbf{RAMA} & \textbf{0.9100} & 0.9100 & \textbf{0.9100} & \textbf{0.9100} & \textbf{0.8200} \\
         \bottomrule
    \end{tabular}
\end{table}

\textbf{Impact of Retrieved Evidence.} To evaluate the importance of external evidence, we removed the retrieved content provided by the WebRetriever module from the input prompts. Both QwenVL and InternVL models exhibited a notable decline in all evaluation metrics in the absence of relevant evidence. This demonstrates that external information is essential for accurate detection, especially for challenging samples where visual-textual cues alone are insufficient. We further illustrate this phenomenon with qualitative examples in Section~\ref{sec: case}.

\textbf{Impact of Agent Ensemble.} We further investigated the effect of agent ensembles by comparing single-agent predictions with those obtained via majority voting across multiple agents and prompt variants. Our results indicate that individual models exhibit complementary strengths: for instance, QwenVL with P1 achieves higher precision, whereas InternVL with P2 yields higher recall. The ensemble approach, which aggregates predictions from diverse models and prompts, consistently achieves superior F1-score and MCC, highlighting the benefit of DecisionFuser for robust detection.

\subsection{Efficiency Analysis}

The efficiency of RAMA is influenced by network conditions and deployment environments. Under the standard experimental configuration described in Section~\ref{sec: imp}, the retrieval module with 7B parameters takes 1 second to collect contextual evidence, while the multimodal validation module with 70B parameters requires 2.5 seconds for multi-agent consistency evaluation, resulting in a total latency of 3.5 seconds. In practical applications, using smaller models may be considered to reduce latency and accommodate resource-constrained environments.

\subsection{Case Study}
\label{sec: case}

This section presents two representative cases in Figure~\ref{fig: case} to illustrate the effectiveness of RAMA framework in multimedia misinformation detection.

\textbf{Case 1: Debunking a Celebrity Death Hoax.}
The first case involves a viral claim regarding celebrity deaths, where visual evidence alone proves insufficient for verification. When presented with an image of Will Smith and his son accompanied by a caption announcing their fatal car accident, RAMA's WebRetriever first establishes an optimal search strategy by analyzing the claim's key components. The system generates the precise query "Will Smith and Jaden Smith died in car crash" and retrieves multiple credible sources, including fact-checking websites and entertainment databases. Crucially, the system identifies a consistent pattern across these sources: while social media platforms propagate the death rumor, authoritative outlets uniformly debunk the claim. This enables RAMA to confidently classify the claim as false, demonstrating its ability to distinguish between viral misinformation and factual reporting through evidence-based reasoning.

\textbf{Case 2: Verifying an Unlikely Real Event.}
The second case presents a more complex scenario where visual evidence appears highly improbable - a metro train seemingly balanced on a whale-shaped art installation. Initial visual-linguistic analysis suggests potential manipulation, as the image violates common physical expectations. However, RAMA's integrated search capability retrieves corroborating reports from multiple reputable news organizations (BBC, NPR, ABC News) that confirm the event's occurrence in Spijkenisse, Netherlands. The system further verifies details about the "Walvisstaarten" sculpture's existence since 2002 and its specific location at the metro line's terminus. This case highlights RAMA's capacity to overcome cognitive biases in visual interpretation by grounding its analysis in retrieved factual evidence, demonstrating its powerful tracing and verification capabilities in dealing with visual forgeries and rare real events.

These case studies highlight RAMA’s robust generalization and verification capabilities, enabling it to effectively discern both fabricated and authentic multimedia content by leveraging retrieved supporting evidence.

\section{Conclusion}
In this work, we proposed RAMA, a retrieval-augmented multi-agent framework for robust multimodal misinformation detection. By integrating advanced visual-language models with a dedicated WebRetriever module, RAMA effectively grounds verification decisions in external, up-to-date evidence. Extensive experiments, including ablation studies and qualitative case analyses, demonstrate that RAMA significantly outperforms text-only and single-agent baselines, particularly in scenarios where visual cues are ambiguous or misleading. Our results highlight the critical importance of evidence retrieval and ensemble reasoning for reliable misinformation detection. Future work will focus on expanding RAMA’s retrieval capabilities to support multilingual and cross-platform evidence, as well as exploring adaptive ensemble strategies to further enhance robustness. We also plan to investigate the framework’s generalization to other domains of multimodal fact-checking. Overall, RAMA offers a promising direction for trustworthy and explainable multimedia verification, bridging the gap between automated misinformation detection and real-world knowledge.

\begin{acks}
This work was supported by Ant Group Research Intern Program.
\end{acks}

\balance
\bibliographystyle{ACM-Reference-Format}
\bibliography{reference}


\begin{thebibliography}{70}


\ifx \showCODEN    \undefined \def \showCODEN     #1{\unskip}     \fi
\ifx \showISBNx    \undefined \def \showISBNx     #1{\unskip}     \fi
\ifx \showISBNxiii \undefined \def \showISBNxiii  #1{\unskip}     \fi
\ifx \showISSN     \undefined \def \showISSN      #1{\unskip}     \fi
\ifx \showLCCN     \undefined \def \showLCCN      #1{\unskip}     \fi
\ifx \shownote     \undefined \def \shownote      #1{#1}          \fi
\ifx \showarticletitle \undefined \def \showarticletitle #1{#1}   \fi
\ifx \showURL      \undefined \def \showURL       {\relax}        \fi
\providecommand\bibfield[2]{#2}
\providecommand\bibinfo[2]{#2}
\providecommand\natexlab[1]{#1}
\providecommand\showeprint[2][]{arXiv:#2}

\bibitem[Abdali et~al\mbox{.}(2024)]%
        {abdali2024multi}
\bibfield{author}{\bibinfo{person}{Sara Abdali}, \bibinfo{person}{Sina Shaham}, {and} \bibinfo{person}{Bhaskar Krishnamachari}.} \bibinfo{year}{2024}\natexlab{}.
\newblock \showarticletitle{Multi-modal misinformation detection: Approaches, challenges and opportunities}.
\newblock \bibinfo{journal}{\emph{Comput. Surveys}} \bibinfo{volume}{57}, \bibinfo{number}{3} (\bibinfo{year}{2024}), \bibinfo{pages}{1--29}.
\newblock


\bibitem[Abdelnabi et~al\mbox{.}(2022)]%
        {abdelnabi2022open}
\bibfield{author}{\bibinfo{person}{Sahar Abdelnabi}, \bibinfo{person}{Rakibul Hasan}, {and} \bibinfo{person}{Mario Fritz}.} \bibinfo{year}{2022}\natexlab{}.
\newblock \showarticletitle{Open-domain, content-based, multi-modal fact-checking of out-of-context images via online resources}. In \bibinfo{booktitle}{\emph{Proceedings of the IEEE/CVF conference on computer vision and pattern recognition}}. \bibinfo{pages}{14940--14949}.
\newblock


\bibitem[Alkaddour et~al\mbox{.}(2022)]%
        {alkaddour2022sentiment}
\bibfield{author}{\bibinfo{person}{Muhannad Alkaddour}, \bibinfo{person}{Abhinav Dhall}, \bibinfo{person}{Usman Tariq}, \bibinfo{person}{Hasan Al~Nashash}, {and} \bibinfo{person}{Fares Al-Shargie}.} \bibinfo{year}{2022}\natexlab{}.
\newblock \showarticletitle{Sentiment-aware classifier for out-of-context caption detection}. In \bibinfo{booktitle}{\emph{Proceedings of the 30th ACM International Conference on Multimedia}}. \bibinfo{pages}{7180--7184}.
\newblock


\bibitem[Aneja et~al\mbox{.}(2023)]%
        {aneja2021cosmos}
\bibfield{author}{\bibinfo{person}{Shivangi Aneja}, \bibinfo{person}{Chris Bregler}, {and} \bibinfo{person}{Matthias Nie\ss{}ner}.} \bibinfo{year}{2023}\natexlab{}.
\newblock \showarticletitle{COSMOS: catching out-of-context image misuse with self-supervised learning}. In \bibinfo{booktitle}{\emph{Proceedings of the Thirty-Seventh AAAI Conference on Artificial Intelligence}} \emph{(\bibinfo{series}{AAAI'23})}. Article \bibinfo{articleno}{1579}, \bibinfo{numpages}{9}~pages.
\newblock
\showISBNx{978-1-57735-880-0}


\bibitem[Apostol et~al\mbox{.}(2024)]%
        {apostol2024contcommrtd}
\bibfield{author}{\bibinfo{person}{Elena-Simona Apostol}, \bibinfo{person}{Ciprian-Octavian Truic{\u{a}}}, {and} \bibinfo{person}{Adrian Paschke}.} \bibinfo{year}{2024}\natexlab{}.
\newblock \showarticletitle{ContCommRTD: A distributed content-based misinformation-aware community detection system for real-time disaster reporting}.
\newblock \bibinfo{journal}{\emph{IEEE Transactions on Knowledge and Data Engineering}} (\bibinfo{year}{2024}).
\newblock


\bibitem[Augenstein et~al\mbox{.}(2024)]%
        {augenstein2024factuality}
\bibfield{author}{\bibinfo{person}{Isabelle Augenstein}, \bibinfo{person}{Timothy Baldwin}, \bibinfo{person}{Meeyoung Cha}, \bibinfo{person}{Tanmoy Chakraborty}, \bibinfo{person}{Giovanni~Luca Ciampaglia}, \bibinfo{person}{David Corney}, \bibinfo{person}{Renee DiResta}, \bibinfo{person}{Emilio Ferrara}, \bibinfo{person}{Scott Hale}, \bibinfo{person}{Alon Halevy}, {et~al\mbox{.}}} \bibinfo{year}{2024}\natexlab{}.
\newblock \showarticletitle{Factuality challenges in the era of large language models and opportunities for fact-checking}.
\newblock \bibinfo{journal}{\emph{Nature Machine Intelligence}} \bibinfo{volume}{6}, \bibinfo{number}{8} (\bibinfo{year}{2024}), \bibinfo{pages}{852--863}.
\newblock


\bibitem[Bai et~al\mbox{.}(2025)]%
        {bai2025qwen2}
\bibfield{author}{\bibinfo{person}{Shuai Bai}, \bibinfo{person}{Keqin Chen}, \bibinfo{person}{Xuejing Liu}, \bibinfo{person}{Jialin Wang}, \bibinfo{person}{Wenbin Ge}, \bibinfo{person}{Sibo Song}, \bibinfo{person}{Kai Dang}, \bibinfo{person}{Peng Wang}, \bibinfo{person}{Shijie Wang}, \bibinfo{person}{Jun Tang}, \bibinfo{person}{Humen Zhong}, \bibinfo{person}{Yuanzhi Zhu}, \bibinfo{person}{Mingkun Yang}, \bibinfo{person}{Zhaohai Li}, \bibinfo{person}{Jianqiang Wan}, \bibinfo{person}{Pengfei Wang}, \bibinfo{person}{Wei Ding}, \bibinfo{person}{Zheren Fu}, \bibinfo{person}{Yiheng Xu}, \bibinfo{person}{Jiabo Ye}, \bibinfo{person}{Xi Zhang}, \bibinfo{person}{Tianbao Xie}, \bibinfo{person}{Zesen Cheng}, \bibinfo{person}{Hang Zhang}, \bibinfo{person}{Zhibo Yang}, \bibinfo{person}{Haiyang Xu}, {and} \bibinfo{person}{Junyang Lin}.} \bibinfo{year}{2025}\natexlab{}.
\newblock \bibinfo{title}{Qwen2.5-VL Technical Report}.
\newblock
\showeprint[arxiv]{2502.13923}


\bibitem[Bondielli and Marcelloni(2019)]%
        {bondielli2019survey}
\bibfield{author}{\bibinfo{person}{Alessandro Bondielli} {and} \bibinfo{person}{Francesco Marcelloni}.} \bibinfo{year}{2019}\natexlab{}.
\newblock \showarticletitle{A survey on fake news and rumour detection techniques}.
\newblock \bibinfo{journal}{\emph{Information sciences}}  \bibinfo{volume}{497} (\bibinfo{year}{2019}), \bibinfo{pages}{38--55}.
\newblock


\bibitem[Botnevik et~al\mbox{.}(2020)]%
        {botnevik2020brenda}
\bibfield{author}{\bibinfo{person}{Bjarte Botnevik}, \bibinfo{person}{Eirik Sakariassen}, {and} \bibinfo{person}{Vinay Setty}.} \bibinfo{year}{2020}\natexlab{}.
\newblock \showarticletitle{Brenda: Browser extension for fake news detection}. In \bibinfo{booktitle}{\emph{Proceedings of the 43rd international ACM SIGIR conference on research and development in information retrieval}}. \bibinfo{pages}{2117--2120}.
\newblock


\bibitem[Cao et~al\mbox{.}(2025)]%
        {cao2025video}
\bibfield{author}{\bibinfo{person}{Meng Cao}, \bibinfo{person}{Pengfei Hu}, \bibinfo{person}{Yingyao Wang}, \bibinfo{person}{Jihao Gu}, \bibinfo{person}{Haoran Tang}, \bibinfo{person}{Haoze Zhao}, \bibinfo{person}{Jiahua Dong}, \bibinfo{person}{Wangbo Yu}, \bibinfo{person}{Ge Zhang}, \bibinfo{person}{Ian Reid}, {et~al\mbox{.}}} \bibinfo{year}{2025}\natexlab{}.
\newblock \bibinfo{title}{Video simpleqa: Towards factuality evaluation in large video language models}.
\newblock
\showeprint[arxiv]{2503.18923}


\bibitem[Cheng et~al\mbox{.}(2020)]%
        {cheng2020vroc}
\bibfield{author}{\bibinfo{person}{Mingxi Cheng}, \bibinfo{person}{Shahin Nazarian}, {and} \bibinfo{person}{Paul Bogdan}.} \bibinfo{year}{2020}\natexlab{}.
\newblock \showarticletitle{Vroc: Variational autoencoder-aided multi-task rumor classifier based on text}. In \bibinfo{booktitle}{\emph{Proceedings of the web conference 2020}}. \bibinfo{pages}{2892--2898}.
\newblock


\bibitem[Choi and Ferrara(2024a)]%
        {choi2024automated}
\bibfield{author}{\bibinfo{person}{Eun~Cheol Choi} {and} \bibinfo{person}{Emilio Ferrara}.} \bibinfo{year}{2024}\natexlab{a}.
\newblock \showarticletitle{Automated claim matching with large language models: empowering fact-checkers in the fight against misinformation}. In \bibinfo{booktitle}{\emph{Companion Proceedings of the ACM Web Conference 2024}}. \bibinfo{pages}{1441--1449}.
\newblock


\bibitem[Choi and Ferrara(2024b)]%
        {choi2024fact}
\bibfield{author}{\bibinfo{person}{Eun~Cheol Choi} {and} \bibinfo{person}{Emilio Ferrara}.} \bibinfo{year}{2024}\natexlab{b}.
\newblock \showarticletitle{Fact-gpt: Fact-checking augmentation via claim matching with llms}. In \bibinfo{booktitle}{\emph{Companion Proceedings of the ACM Web Conference 2024}}. \bibinfo{pages}{883--886}.
\newblock


\bibitem[Dang-Nguyen et~al\mbox{.}(2023)]%
        {dang2023grand}
\bibfield{author}{\bibinfo{person}{Duc-Tien Dang-Nguyen}, \bibinfo{person}{Sohail~Ahmed Khan}, \bibinfo{person}{Cise Midoglu}, \bibinfo{person}{Michael Riegler}, \bibinfo{person}{P{\aa}l Halvorsen}, {and} \bibinfo{person}{Minh-Son Dao}.} \bibinfo{year}{2023}\natexlab{}.
\newblock \bibinfo{title}{Grand challenge on detecting cheapfakes}.
\newblock
\showeprint[arxiv]{2304.01328}


\bibitem[Dang-Nguyen et~al\mbox{.}(2024)]%
        {nguyen2024overview}
\bibfield{author}{\bibinfo{person}{Duc-Tien Dang-Nguyen}, \bibinfo{person}{Sohail~Ahmed Khan}, \bibinfo{person}{Michael Riegler}, \bibinfo{person}{P{\aa}l Halvorsen}, \bibinfo{person}{Anh-Duy Tran}, \bibinfo{person}{Minh-Son Dao}, {and} \bibinfo{person}{Minh-Triet Tran}.} \bibinfo{year}{2024}\natexlab{}.
\newblock \showarticletitle{Overview of the grand challenge on detecting cheapfakes at ACM ICMR 2024}. In \bibinfo{booktitle}{\emph{Proceedings of the 2024 International Conference on Multimedia Retrieval}}. \bibinfo{pages}{1275--1281}.
\newblock


\bibitem[Dang-Nguyen et~al\mbox{.}(2025)]%
        {MV2025overview}
\bibfield{author}{\bibinfo{person}{Duc-Tien Dang-Nguyen}, \bibinfo{person}{Morten Langfeldt~Dahlback}, \bibinfo{person}{Henrik Vold}, \bibinfo{person}{Silje Førsund}, \bibinfo{person}{Minh-Son Dao}, \bibinfo{person}{Kha-Luan Pham}, \bibinfo{person}{Sohail~Ahmed Khan}, \bibinfo{person}{Marc Gallofré~Ocaña}, \bibinfo{person}{Minh-Triet Tran}, {and} \bibinfo{person}{Anh-Duy Tran}.} \bibinfo{year}{2025}\natexlab{}.
\newblock \showarticletitle{The 2025 Grand Challenge on Multimedia Verification: Foundations and Overview}. In \bibinfo{booktitle}{\emph{Proceedings of the 33rd ACM International Conference on Multimedia}}.
\newblock


\bibitem[Guo et~al\mbox{.}(2022)]%
        {guo2022survey}
\bibfield{author}{\bibinfo{person}{Zhijiang Guo}, \bibinfo{person}{Michael Schlichtkrull}, {and} \bibinfo{person}{Andreas Vlachos}.} \bibinfo{year}{2022}\natexlab{}.
\newblock \showarticletitle{A survey on automated fact-checking}.
\newblock \bibinfo{journal}{\emph{Transactions of the Association for Computational Linguistics}}  \bibinfo{volume}{10} (\bibinfo{year}{2022}), \bibinfo{pages}{178--206}.
\newblock


\bibitem[Huan et~al\mbox{.}(2024)]%
        {huan2024social}
\bibfield{author}{\bibinfo{person}{Xing Huan}, \bibinfo{person}{Antonio Parbonetti}, \bibinfo{person}{Giulia Redigolo}, {and} \bibinfo{person}{Zhewei Zhang}.} \bibinfo{year}{2024}\natexlab{}.
\newblock \showarticletitle{Social media disclosure and reputational damage}.
\newblock \bibinfo{journal}{\emph{Review of Quantitative Finance and Accounting}} \bibinfo{volume}{62}, \bibinfo{number}{4} (\bibinfo{year}{2024}), \bibinfo{pages}{1355--1396}.
\newblock


\bibitem[Huang et~al\mbox{.}(2020)]%
        {huang2020conquering}
\bibfield{author}{\bibinfo{person}{Yen-Hao Huang}, \bibinfo{person}{Ting-Wei Liu}, \bibinfo{person}{Ssu-Rui Lee}, \bibinfo{person}{Fernando~Henrique Calderon~Alvarado}, {and} \bibinfo{person}{Yi-Shin Chen}.} \bibinfo{year}{2020}\natexlab{}.
\newblock \showarticletitle{Conquering cross-source failure for news credibility: Learning generalizable representations beyond content embedding}. In \bibinfo{booktitle}{\emph{Proceedings of The Web Conference 2020}}. \bibinfo{pages}{774--784}.
\newblock


\bibitem[Islam et~al\mbox{.}(2020)]%
        {islam2020deep}
\bibfield{author}{\bibinfo{person}{Md~Rafiqul Islam}, \bibinfo{person}{Shaowu Liu}, \bibinfo{person}{Xianzhi Wang}, {and} \bibinfo{person}{Guandong Xu}.} \bibinfo{year}{2020}\natexlab{}.
\newblock \showarticletitle{Deep learning for misinformation detection on online social networks: a survey and new perspectives}.
\newblock \bibinfo{journal}{\emph{Social Network Analysis and Mining}} \bibinfo{volume}{10}, \bibinfo{number}{1} (\bibinfo{year}{2020}), \bibinfo{pages}{82}.
\newblock


\bibitem[Ito et~al\mbox{.}(2015)]%
        {ito2015assessment}
\bibfield{author}{\bibinfo{person}{Jun Ito}, \bibinfo{person}{Jing Song}, \bibinfo{person}{Hiroyuki Toda}, \bibinfo{person}{Yoshimasa Koike}, {and} \bibinfo{person}{Satoshi Oyama}.} \bibinfo{year}{2015}\natexlab{}.
\newblock \showarticletitle{Assessment of tweet credibility with LDA features}. In \bibinfo{booktitle}{\emph{Proceedings of the 24th international conference on world wide web}}. \bibinfo{pages}{953--958}.
\newblock


\bibitem[Jaiswal et~al\mbox{.}(2019)]%
        {jaiswal2019aird}
\bibfield{author}{\bibinfo{person}{Ayush Jaiswal}, \bibinfo{person}{Yue Wu}, \bibinfo{person}{Wael AbdAlmageed}, \bibinfo{person}{Iacopo Masi}, {and} \bibinfo{person}{Premkumar Natarajan}.} \bibinfo{year}{2019}\natexlab{}.
\newblock \showarticletitle{Aird: Adversarial learning framework for image repurposing detection}. In \bibinfo{booktitle}{\emph{Proceedings of the IEEE/CVF Conference on Computer Vision and Pattern Recognition}}. \bibinfo{pages}{11330--11339}.
\newblock


\bibitem[Jia et~al\mbox{.}(2024)]%
        {jia2024can}
\bibfield{author}{\bibinfo{person}{Shan Jia}, \bibinfo{person}{Reilin Lyu}, \bibinfo{person}{Kangran Zhao}, \bibinfo{person}{Yize Chen}, \bibinfo{person}{Zhiyuan Yan}, \bibinfo{person}{Yan Ju}, \bibinfo{person}{Chuanbo Hu}, \bibinfo{person}{Xin Li}, \bibinfo{person}{Baoyuan Wu}, {and} \bibinfo{person}{Siwei Lyu}.} \bibinfo{year}{2024}\natexlab{}.
\newblock \showarticletitle{Can chatgpt detect deepfakes? a study of using multimodal large language models for media forensics}. In \bibinfo{booktitle}{\emph{Proceedings of the IEEE/CVF Conference on Computer Vision and Pattern Recognition}}. \bibinfo{pages}{4324--4333}.
\newblock


\bibitem[Khattar et~al\mbox{.}(2019)]%
        {khattar2019mvae}
\bibfield{author}{\bibinfo{person}{Dhruv Khattar}, \bibinfo{person}{Jaipal~Singh Goud}, \bibinfo{person}{Manish Gupta}, {and} \bibinfo{person}{Vasudeva Varma}.} \bibinfo{year}{2019}\natexlab{}.
\newblock \showarticletitle{Mvae: Multimodal variational autoencoder for fake news detection}. In \bibinfo{booktitle}{\emph{The world wide web conference}}. \bibinfo{pages}{2915--2921}.
\newblock


\bibitem[Kumari et~al\mbox{.}(2021)]%
        {kumari2021misinformation}
\bibfield{author}{\bibinfo{person}{Rina Kumari}, \bibinfo{person}{Nischal Ashok}, \bibinfo{person}{Tirthankar Ghosal}, {and} \bibinfo{person}{Asif Ekbal}.} \bibinfo{year}{2021}\natexlab{}.
\newblock \showarticletitle{Misinformation detection using multitask learning with mutual learning for novelty detection and emotion recognition}.
\newblock \bibinfo{journal}{\emph{Information Processing \& Management}} \bibinfo{volume}{58}, \bibinfo{number}{5} (\bibinfo{year}{2021}), \bibinfo{pages}{102631}.
\newblock


\bibitem[Le et~al\mbox{.}(2024)]%
        {le2024tega}
\bibfield{author}{\bibinfo{person}{Anh-Thu Le}, \bibinfo{person}{Minh-Dat Nguyen}, \bibinfo{person}{Minh-Son Dao}, \bibinfo{person}{Anh-Duy Tran}, {and} \bibinfo{person}{Duc-Tien Dang-Nguyen}.} \bibinfo{year}{2024}\natexlab{}.
\newblock \showarticletitle{TeGA: A Text-Guided Generative-based Approach in Cheapfake Detection}. In \bibinfo{booktitle}{\emph{Proceedings of the 2024 International Conference on Multimedia Retrieval}}. \bibinfo{pages}{1294--1299}.
\newblock


\bibitem[Li et~al\mbox{.}(2024)]%
        {li2024large}
\bibfield{author}{\bibinfo{person}{Xinyi Li}, \bibinfo{person}{Yongfeng Zhang}, {and} \bibinfo{person}{Edward~C Malthouse}.} \bibinfo{year}{2024}\natexlab{}.
\newblock \bibinfo{title}{Large language model agent for fake news detection}.
\newblock
\showeprint[arxiv]{2405.01593}


\bibitem[Luo et~al\mbox{.}(2021)]%
        {luo2021newsclippings}
\bibfield{author}{\bibinfo{person}{Grace Luo}, \bibinfo{person}{Trevor Darrell}, {and} \bibinfo{person}{Anna Rohrbach}.} \bibinfo{year}{2021}\natexlab{}.
\newblock \showarticletitle{NewsCLIPpings: Automatic Generation of Out-of-Context Multimodal Media}. In \bibinfo{booktitle}{\emph{Proceedings of the 2021 Conference on Empirical Methods in Natural Language Processing}}. \bibinfo{pages}{6801--6817}.
\newblock


\bibitem[Ma et~al\mbox{.}(2025)]%
        {ma2025local}
\bibfield{author}{\bibinfo{person}{Jiatong Ma}, \bibinfo{person}{Linmei Hu}, \bibinfo{person}{Rang Li}, {and} \bibinfo{person}{Wenbo Fu}.} \bibinfo{year}{2025}\natexlab{}.
\newblock \showarticletitle{LoCal: Logical and Causal Fact-Checking with LLM-Based Multi-Agents}. In \bibinfo{booktitle}{\emph{Proceedings of the ACM on Web Conference 2025}}. \bibinfo{pages}{1614--1625}.
\newblock


\bibitem[Masood et~al\mbox{.}(2023)]%
        {masood2023deepfakes}
\bibfield{author}{\bibinfo{person}{Momina Masood}, \bibinfo{person}{Mariam Nawaz}, \bibinfo{person}{Khalid~Mahmood Malik}, \bibinfo{person}{Ali Javed}, \bibinfo{person}{Aun Irtaza}, {and} \bibinfo{person}{Hafiz Malik}.} \bibinfo{year}{2023}\natexlab{}.
\newblock \showarticletitle{Deepfakes generation and detection: State-of-the-art, open challenges, countermeasures, and way forward}.
\newblock \bibinfo{journal}{\emph{Applied intelligence}} \bibinfo{volume}{53}, \bibinfo{number}{4} (\bibinfo{year}{2023}), \bibinfo{pages}{3974--4026}.
\newblock


\bibitem[Meel and Vishwakarma(2020)]%
        {meel2020fake}
\bibfield{author}{\bibinfo{person}{Priyanka Meel} {and} \bibinfo{person}{Dinesh~Kumar Vishwakarma}.} \bibinfo{year}{2020}\natexlab{}.
\newblock \showarticletitle{Fake news, rumor, information pollution in social media and web: A contemporary survey of state-of-the-arts, challenges and opportunities}.
\newblock \bibinfo{journal}{\emph{Expert Systems with Applications}}  \bibinfo{volume}{153} (\bibinfo{year}{2020}), \bibinfo{pages}{112986}.
\newblock


\bibitem[Micallef et~al\mbox{.}(2022)]%
        {micallef2022cross}
\bibfield{author}{\bibinfo{person}{Nicholas Micallef}, \bibinfo{person}{Marcelo Sandoval-Casta{\~n}eda}, \bibinfo{person}{Adi Cohen}, \bibinfo{person}{Mustaque Ahamad}, \bibinfo{person}{Srijan Kumar}, {and} \bibinfo{person}{Nasir Memon}.} \bibinfo{year}{2022}\natexlab{}.
\newblock \showarticletitle{Cross-platform multimodal misinformation: taxonomy, characteristics and detection for textual posts and videos}. In \bibinfo{booktitle}{\emph{Proceedings of the International AAAI Conference on Web and Social Media}}, Vol.~\bibinfo{volume}{16}. \bibinfo{pages}{651--662}.
\newblock


\bibitem[Mu et~al\mbox{.}(2023)]%
        {mu2023self}
\bibfield{author}{\bibinfo{person}{Michael Mu}, \bibinfo{person}{Sreyasee Das~Bhattacharjee}, {and} \bibinfo{person}{Junsong Yuan}.} \bibinfo{year}{2023}\natexlab{}.
\newblock \showarticletitle{Self-supervised distilled learning for multi-modal misinformation identification}. In \bibinfo{booktitle}{\emph{Proceedings of the IEEE/CVF Winter Conference on Applications of Computer Vision}}. \bibinfo{pages}{2819--2828}.
\newblock


\bibitem[Nguyen et~al\mbox{.}(2024)]%
        {nguyen2024unified}
\bibfield{author}{\bibinfo{person}{Van-Loc Nguyen}, \bibinfo{person}{Bao-Tin Nguyen}, \bibinfo{person}{Thanh-Son Nguyen}, \bibinfo{person}{Duc-Tien Dang-Nguyen}, {and} \bibinfo{person}{Minh-Triet Tran}.} \bibinfo{year}{2024}\natexlab{}.
\newblock \showarticletitle{A Unified Network for Detecting Out-Of-Context Information Using Generative Synthetic Data}. In \bibinfo{booktitle}{\emph{Proceedings of the 2024 International Conference on Multimedia Retrieval}}. \bibinfo{pages}{1300--1305}.
\newblock


\bibitem[Okati-Aliabad et~al\mbox{.}(2024)]%
        {okati2024truth}
\bibfield{author}{\bibinfo{person}{Hassan Okati-Aliabad}, \bibinfo{person}{Mahdi Mohammadi}, \bibinfo{person}{Alireza~Salimi Khorashad}, \bibinfo{person}{Alireza Ansari-Moghaddam}, \bibinfo{person}{Mohsen~Hossein Bor}, {and} \bibinfo{person}{Jalil Nejati}.} \bibinfo{year}{2024}\natexlab{}.
\newblock \showarticletitle{A Truth or a Rumor: Effects of Addictive Substances on Prevention of COVID-19; an Investigation of Homeless Drug Abusers in Southeastern Iran}.
\newblock \bibinfo{journal}{\emph{International Journal of High Risk Behaviors and Addiction}} \bibinfo{volume}{13}, \bibinfo{number}{1} (\bibinfo{year}{2024}).
\newblock


\bibitem[Papadopoulos et~al\mbox{.}(2025a)]%
        {papadopoulos2025red}
\bibfield{author}{\bibinfo{person}{Stefanos-Iordanis Papadopoulos}, \bibinfo{person}{Christos Koutlis}, \bibinfo{person}{Symeon Papadopoulos}, {and} \bibinfo{person}{Panagiotis~C Petrantonakis}.} \bibinfo{year}{2025}\natexlab{a}.
\newblock \showarticletitle{Red-dot: Multimodal fact-checking via relevant evidence detection}.
\newblock \bibinfo{journal}{\emph{IEEE Transactions on Computational Social Systems}} (\bibinfo{year}{2025}).
\newblock


\bibitem[Papadopoulos et~al\mbox{.}(2025b)]%
        {papadopoulos2025similarity}
\bibfield{author}{\bibinfo{person}{Stefanos-Iordanis Papadopoulos}, \bibinfo{person}{Christos Koutlis}, \bibinfo{person}{Symeon Papadopoulos}, {and} \bibinfo{person}{Panagiotis~C Petrantonakis}.} \bibinfo{year}{2025}\natexlab{b}.
\newblock \showarticletitle{Similarity over Factuality: Are we making progress on multimodal out-of-context misinformation detection?}. In \bibinfo{booktitle}{\emph{2025 IEEE/CVF Winter Conference on Applications of Computer Vision (WACV)}}. IEEE, \bibinfo{pages}{5041--5050}.
\newblock


\bibitem[Paris and Donovan(2019)]%
        {paris2019deepfakes}
\bibfield{author}{\bibinfo{person}{Britt Paris} {and} \bibinfo{person}{Joan Donovan}.} \bibinfo{year}{2019}\natexlab{}.
\newblock \showarticletitle{Deepfakes and cheap fakes}.
\newblock \bibinfo{journal}{\emph{United States of America: Data \& Society}}  \bibinfo{volume}{1} (\bibinfo{year}{2019}).
\newblock


\bibitem[Pelrine et~al\mbox{.}(2021)]%
        {pelrine2021surprising}
\bibfield{author}{\bibinfo{person}{Kellin Pelrine}, \bibinfo{person}{Jacob Danovitch}, {and} \bibinfo{person}{Reihaneh Rabbany}.} \bibinfo{year}{2021}\natexlab{}.
\newblock \showarticletitle{The surprising performance of simple baselines for misinformation detection}. In \bibinfo{booktitle}{\emph{Proceedings of the Web Conference 2021}}. \bibinfo{pages}{3432--3441}.
\newblock


\bibitem[Pham et~al\mbox{.}(2024)]%
        {pham2024generative}
\bibfield{author}{\bibinfo{person}{Long-Khanh Pham}, \bibinfo{person}{Hoa-Vien Vo-Hoang}, {and} \bibinfo{person}{Anh-Duy Tran}.} \bibinfo{year}{2024}\natexlab{}.
\newblock \showarticletitle{A Generative Adaptive Context Learning Framework for Large Language Models in Cheapfake Detection}. In \bibinfo{booktitle}{\emph{Proceedings of the 2024 International Conference on Multimedia Retrieval}}. \bibinfo{pages}{1288--1293}.
\newblock


\bibitem[Qi et~al\mbox{.}(2024)]%
        {qi2024sniffer}
\bibfield{author}{\bibinfo{person}{Peng Qi}, \bibinfo{person}{Zehong Yan}, \bibinfo{person}{Wynne Hsu}, {and} \bibinfo{person}{Mong~Li Lee}.} \bibinfo{year}{2024}\natexlab{}.
\newblock \showarticletitle{Sniffer: Multimodal large language model for explainable out-of-context misinformation detection}. In \bibinfo{booktitle}{\emph{Proceedings of the IEEE/CVF conference on computer vision and pattern recognition}}. \bibinfo{pages}{13052--13062}.
\newblock


\bibitem[Qian et~al\mbox{.}(2018)]%
        {qian2018neural}
\bibfield{author}{\bibinfo{person}{Feng Qian}, \bibinfo{person}{Chengyue Gong}, \bibinfo{person}{Karishma Sharma}, {and} \bibinfo{person}{Yan Liu}.} \bibinfo{year}{2018}\natexlab{}.
\newblock \showarticletitle{Neural user response generator: fake news detection with collective user intelligence}. In \bibinfo{booktitle}{\emph{Proceedings of the 27th International Joint Conference on Artificial Intelligence}}. \bibinfo{pages}{3834--3840}.
\newblock


\bibitem[Qian et~al\mbox{.}(2021)]%
        {qian2021knowledge}
\bibfield{author}{\bibinfo{person}{Shengsheng Qian}, \bibinfo{person}{Jun Hu}, \bibinfo{person}{Quan Fang}, {and} \bibinfo{person}{Changsheng Xu}.} \bibinfo{year}{2021}\natexlab{}.
\newblock \showarticletitle{Knowledge-aware multi-modal adaptive graph convolutional networks for fake news detection}.
\newblock \bibinfo{journal}{\emph{ACM Transactions on Multimedia Computing, Communications, and Applications (TOMM)}} \bibinfo{volume}{17}, \bibinfo{number}{3} (\bibinfo{year}{2021}), \bibinfo{pages}{1--23}.
\newblock


\bibitem[Radford et~al\mbox{.}(2021)]%
        {radford2021learningtransferablevisualmodels}
\bibfield{author}{\bibinfo{person}{Alec Radford}, \bibinfo{person}{Jong~Wook Kim}, \bibinfo{person}{Chris Hallacy}, \bibinfo{person}{Aditya Ramesh}, \bibinfo{person}{Gabriel Goh}, \bibinfo{person}{Sandhini Agarwal}, \bibinfo{person}{Girish Sastry}, \bibinfo{person}{Amanda Askell}, \bibinfo{person}{Pamela Mishkin}, \bibinfo{person}{Jack Clark}, \bibinfo{person}{Gretchen Krueger}, {and} \bibinfo{person}{Ilya Sutskever}.} \bibinfo{year}{2021}\natexlab{}.
\newblock \bibinfo{title}{Learning Transferable Visual Models From Natural Language Supervision}.
\newblock
\showeprint[arxiv]{2103.00020}


\bibitem[Rayar et~al\mbox{.}(2022)]%
        {rayar2022large}
\bibfield{author}{\bibinfo{person}{Fr{\'e}d{\'e}ric Rayar}, \bibinfo{person}{Mathieu Delalandre}, {and} \bibinfo{person}{Van-Hao Le}.} \bibinfo{year}{2022}\natexlab{}.
\newblock \showarticletitle{A large-scale TV video and metadata database for French political content analysis and fact-checking}. In \bibinfo{booktitle}{\emph{Proceedings of the 19th International Conference on Content-based Multimedia Indexing}}. \bibinfo{pages}{181--185}.
\newblock


\bibitem[Saakyan et~al\mbox{.}(2021)]%
        {Saakyan2021}
\bibfield{author}{\bibinfo{person}{Aram Saakyan}, \bibinfo{person}{Bharathi~Raja Chakravarthi}, {and} \bibinfo{person}{et al.}} \bibinfo{year}{2021}\natexlab{}.
\newblock \showarticletitle{COVID-fact: Fact Extraction and Verification of Real-World Claims on COVID-19 Pandemic}. In \bibinfo{booktitle}{\emph{Proceedings of ACL 2021}}.
\newblock


\bibitem[Seo et~al\mbox{.}(2024)]%
        {seo2024multi}
\bibfield{author}{\bibinfo{person}{Jangwon Seo}, \bibinfo{person}{Hyo-Seok Hwang}, \bibinfo{person}{Jiyoung Lee}, \bibinfo{person}{Minhyeok Lee}, \bibinfo{person}{Wonsuk Kim}, {and} \bibinfo{person}{Junhee Seok}.} \bibinfo{year}{2024}\natexlab{}.
\newblock \showarticletitle{A Multi-Stage Deep Learning Approach Incorporating Text-Image and Image-Image Comparisons for Cheapfake Detection}. In \bibinfo{booktitle}{\emph{Proceedings of the 2024 International Conference on Multimedia Retrieval}}. \bibinfo{pages}{1312--1316}.
\newblock


\bibitem[Shang et~al\mbox{.}(2024)]%
        {shang2024mmadapt}
\bibfield{author}{\bibinfo{person}{Lanyu Shang}, \bibinfo{person}{Yang Zhang}, \bibinfo{person}{Bozhang Chen}, \bibinfo{person}{Ruohan Zong}, \bibinfo{person}{Zhenrui Yue}, \bibinfo{person}{Huimin Zeng}, \bibinfo{person}{Na Wei}, {and} \bibinfo{person}{Dong Wang}.} \bibinfo{year}{2024}\natexlab{}.
\newblock \showarticletitle{MMAdapt: A Knowledge-guided Multi-source Multi-class Domain Adaptive Framework for Early Health Misinformation Detection}. In \bibinfo{booktitle}{\emph{Proceedings of the ACM Web Conference 2024}}. \bibinfo{pages}{4653--4663}.
\newblock


\bibitem[Shu et~al\mbox{.}(2019)]%
        {shu2019defend}
\bibfield{author}{\bibinfo{person}{Kai Shu}, \bibinfo{person}{Limeng Cui}, \bibinfo{person}{Suhang Wang}, \bibinfo{person}{Dongwon Lee}, {and} \bibinfo{person}{Huan Liu}.} \bibinfo{year}{2019}\natexlab{}.
\newblock \showarticletitle{defend: Explainable fake news detection}. In \bibinfo{booktitle}{\emph{Proceedings of the 25th ACM SIGKDD international conference on knowledge discovery \& data mining}}. \bibinfo{pages}{395--405}.
\newblock


\bibitem[Tang et~al\mbox{.}(2024)]%
        {tang2024minicheck}
\bibfield{author}{\bibinfo{person}{Liyan Tang}, \bibinfo{person}{Philippe Laban}, {and} \bibinfo{person}{Greg Durrett}.} \bibinfo{year}{2024}\natexlab{}.
\newblock \showarticletitle{MiniCheck: Efficient Fact-Checking of LLMs on Grounding Documents}. In \bibinfo{booktitle}{\emph{Proceedings of the 2024 Conference on Empirical Methods in Natural Language Processing}}. \bibinfo{pages}{8818--8847}.
\newblock


\bibitem[Tran et~al\mbox{.}(2022)]%
        {tran2022textual}
\bibfield{author}{\bibinfo{person}{Quang-Tien Tran}, \bibinfo{person}{Thanh-Phuc Tran}, \bibinfo{person}{Minh-Son Dao}, \bibinfo{person}{Tuan-Vinh La}, \bibinfo{person}{Anh-Duy Tran}, {and} \bibinfo{person}{Duc~Tien Dang~Nguyen}.} \bibinfo{year}{2022}\natexlab{}.
\newblock \showarticletitle{A textual-visual-entailment-based unsupervised algorithm for cheapfake detection}. In \bibinfo{booktitle}{\emph{Proceedings of the 30th ACM International Conference on Multimedia}}. \bibinfo{pages}{7145--7149}.
\newblock


\bibitem[Vo-Hoang et~al\mbox{.}(2024)]%
        {vo2024detecting}
\bibfield{author}{\bibinfo{person}{Hoa-Vien Vo-Hoang}, \bibinfo{person}{Long-Khanh Pham}, {and} \bibinfo{person}{Minh-Son Dao}.} \bibinfo{year}{2024}\natexlab{}.
\newblock \showarticletitle{Detecting out-of-context media with llama-adapter v2 and roberta: An effective method for cheapfakes detection}. In \bibinfo{booktitle}{\emph{Proceedings of the 2024 International Conference on Multimedia Retrieval}}. \bibinfo{pages}{1282--1287}.
\newblock


\bibitem[Vu et~al\mbox{.}(2024)]%
        {vu2024enhancing}
\bibfield{author}{\bibinfo{person}{Dang Vu}, \bibinfo{person}{Minh-Nhat Nguyen}, {and} \bibinfo{person}{Quoc-Trung Nguyen}.} \bibinfo{year}{2024}\natexlab{}.
\newblock \showarticletitle{Enhancing Cheapfake Detection: An Approach Using Prompt Engineering and Interleaved Text-Image Model}. In \bibinfo{booktitle}{\emph{Proceedings of the 2024 International Conference on Multimedia Retrieval}}. \bibinfo{pages}{1306--1311}.
\newblock


\bibitem[Wadden et~al\mbox{.}(2020)]%
        {Wadden2020}
\bibfield{author}{\bibinfo{person}{David Wadden}, \bibinfo{person}{Shanchan Lin}, \bibinfo{person}{Kyle Lo}, {and} \bibinfo{person}{et al.}} \bibinfo{year}{2020}\natexlab{}.
\newblock \showarticletitle{Fact or Fiction: Verifying Scientific Claims}. In \bibinfo{booktitle}{\emph{EMNLP 2020}}.
\newblock


\bibitem[Wan et~al\mbox{.}(2024)]%
        {wan2024dell}
\bibfield{author}{\bibinfo{person}{Herun Wan}, \bibinfo{person}{Shangbin Feng}, \bibinfo{person}{Zhaoxuan Tan}, \bibinfo{person}{Heng Wang}, \bibinfo{person}{Yulia Tsvetkov}, {and} \bibinfo{person}{Minnan Luo}.} \bibinfo{year}{2024}\natexlab{}.
\newblock \showarticletitle{DELL: Generating Reactions and Explanations for LLM-Based Misinformation Detection}. In \bibinfo{booktitle}{\emph{Findings of the Association for Computational Linguistics ACL 2024}}. \bibinfo{pages}{2637--2667}.
\newblock


\bibitem[Wang et~al\mbox{.}(2025b)]%
        {wang2025decoupling}
\bibfield{author}{\bibinfo{person}{Guoqing Wang}, \bibinfo{person}{Wen Wu}, \bibinfo{person}{Guangze Ye}, \bibinfo{person}{Zhenxiao Cheng}, \bibinfo{person}{Xi Chen}, {and} \bibinfo{person}{Hong Zheng}.} \bibinfo{year}{2025}\natexlab{b}.
\newblock \showarticletitle{Decoupling Metacognition from Cognition: A Framework for Quantifying Metacognitive Ability in LLMs}. In \bibinfo{booktitle}{\emph{Proceedings of the AAAI Conference on Artificial Intelligence}}, Vol.~\bibinfo{volume}{39}. \bibinfo{pages}{25353--25361}.
\newblock


\bibitem[Wang et~al\mbox{.}(2025a)]%
        {wang2024mfc}
\bibfield{author}{\bibinfo{person}{Shengkang Wang}, \bibinfo{person}{Hongzhan Lin}, \bibinfo{person}{Ziyang Luo}, \bibinfo{person}{Zhen Ye}, \bibinfo{person}{Guang Chen}, {and} \bibinfo{person}{Jing Ma}.} \bibinfo{year}{2025}\natexlab{a}.
\newblock \bibinfo{title}{Mfc-bench: Benchmarking multimodal fact-checking with large vision-language models}.
\newblock
\showeprint[arxiv]{2406.11288}


\bibitem[Warren et~al\mbox{.}(2025)]%
        {warren2025show}
\bibfield{author}{\bibinfo{person}{Greta Warren}, \bibinfo{person}{Irina Shklovski}, {and} \bibinfo{person}{Isabelle Augenstein}.} \bibinfo{year}{2025}\natexlab{}.
\newblock \showarticletitle{Show Me the Work: Fact-Checkers' Requirements for Explainable Automated Fact-Checking}. In \bibinfo{booktitle}{\emph{Proceedings of the 2025 CHI Conference on Human Factors in Computing Systems}}. \bibinfo{pages}{1--21}.
\newblock


\bibitem[Wu et~al\mbox{.}(2023)]%
        {wu2023cheap}
\bibfield{author}{\bibinfo{person}{Guangyang Wu}, \bibinfo{person}{Weijie Wu}, \bibinfo{person}{Xiaohong Liu}, \bibinfo{person}{Kele Xu}, \bibinfo{person}{Tianjiao Wan}, {and} \bibinfo{person}{Wenyi Wang}.} \bibinfo{year}{2023}\natexlab{}.
\newblock \showarticletitle{Cheap-fake detection with llm using prompt engineering}. In \bibinfo{booktitle}{\emph{2023 IEEE International Conference on Multimedia and Expo Workshops (ICMEW)}}. IEEE, \bibinfo{pages}{105--109}.
\newblock


\bibitem[Xu et~al\mbox{.}(2024)]%
        {xu2024mmooc}
\bibfield{author}{\bibinfo{person}{Qingzheng Xu}, \bibinfo{person}{Heming Du}, \bibinfo{person}{Huiqiang Chen}, \bibinfo{person}{Bo Liu}, {and} \bibinfo{person}{Xin Yu}.} \bibinfo{year}{2024}\natexlab{}.
\newblock \showarticletitle{MMOOC: A Multimodal Misinformation Dataset for Out-of-Context News Analysis}. In \bibinfo{booktitle}{\emph{Australasian Conference on Information Security and Privacy}}. Springer, \bibinfo{pages}{444--459}.
\newblock


\bibitem[Yang et~al\mbox{.}(2025a)]%
        {yang2025realfactbench}
\bibfield{author}{\bibinfo{person}{Shuo Yang}, \bibinfo{person}{Yuqin Dai}, \bibinfo{person}{Guoqing Wang}, \bibinfo{person}{Xinran Zheng}, \bibinfo{person}{Jinfeng Xu}, \bibinfo{person}{Jinze Li}, \bibinfo{person}{Zhenzhe Ying}, \bibinfo{person}{Weiqiang Wang}, {and} \bibinfo{person}{Edith C.~H. Ngai}.} \bibinfo{year}{2025}\natexlab{a}.
\newblock \bibinfo{title}{RealFactBench: A Benchmark for Evaluating Large Language Models in Real-World Fact-Checking}.
\newblock
\showeprint[arxiv]{2506.12538}


\bibitem[Yang et~al\mbox{.}(2025b)]%
        {yang2025largelanguagemodelsnetwork}
\bibfield{author}{\bibinfo{person}{Shuo Yang}, \bibinfo{person}{Xinran Zheng}, \bibinfo{person}{Xinchen Zhang}, \bibinfo{person}{Jinfeng Xu}, \bibinfo{person}{Jinze Li}, \bibinfo{person}{Donglin Xie}, \bibinfo{person}{Weicai Long}, {and} \bibinfo{person}{Edith C.~H. Ngai}.} \bibinfo{year}{2025}\natexlab{b}.
\newblock \bibinfo{title}{Large Language Models for Network Intrusion Detection Systems: Foundations, Implementations, and Future Directions}.
\newblock
\showeprint[arxiv]{2507.04752}


\bibitem[Yao et~al\mbox{.}(2023)]%
        {yao2023end}
\bibfield{author}{\bibinfo{person}{Barry~Menglong Yao}, \bibinfo{person}{Aditya Shah}, \bibinfo{person}{Lichao Sun}, \bibinfo{person}{Jin-Hee Cho}, {and} \bibinfo{person}{Lifu Huang}.} \bibinfo{year}{2023}\natexlab{}.
\newblock \showarticletitle{End-to-end multimodal fact-checking and explanation generation: A challenging dataset and models}. In \bibinfo{booktitle}{\emph{Proceedings of the 46th International ACM SIGIR Conference on Research and Development in Information Retrieval}}. \bibinfo{pages}{2733--2743}.
\newblock


\bibitem[Yu et~al\mbox{.}(2017)]%
        {yu2017convolutional}
\bibfield{author}{\bibinfo{person}{Feng Yu}, \bibinfo{person}{Qiang Liu}, \bibinfo{person}{Shu Wu}, \bibinfo{person}{Liang Wang}, \bibinfo{person}{Tieniu Tan}, {et~al\mbox{.}}} \bibinfo{year}{2017}\natexlab{}.
\newblock \showarticletitle{A convolutional approach for misinformation identification.}. In \bibinfo{booktitle}{\emph{IJCAI}}, Vol.~\bibinfo{volume}{2017}. \bibinfo{pages}{3901--3907}.
\newblock


\bibitem[Zhang et~al\mbox{.}(2016)]%
        {zhang2016misinformation}
\bibfield{author}{\bibinfo{person}{Huiling Zhang}, \bibinfo{person}{Md~Abdul Alim}, \bibinfo{person}{Xiang Li}, \bibinfo{person}{My~T Thai}, {and} \bibinfo{person}{Hien~T Nguyen}.} \bibinfo{year}{2016}\natexlab{}.
\newblock \showarticletitle{Misinformation in online social networks: Detect them all with a limited budget}.
\newblock \bibinfo{journal}{\emph{ACM Transactions on Information Systems (TOIS)}} \bibinfo{volume}{34}, \bibinfo{number}{3} (\bibinfo{year}{2016}), \bibinfo{pages}{1--24}.
\newblock


\bibitem[Zhang et~al\mbox{.}(2024)]%
        {zhang2024toward}
\bibfield{author}{\bibinfo{person}{Yizhou Zhang}, \bibinfo{person}{Karishma Sharma}, \bibinfo{person}{Lun Du}, {and} \bibinfo{person}{Yan Liu}.} \bibinfo{year}{2024}\natexlab{}.
\newblock \showarticletitle{Toward mitigating misinformation and social media manipulation in llm era}. In \bibinfo{booktitle}{\emph{Companion Proceedings of the ACM Web Conference 2024}}. \bibinfo{pages}{1302--1305}.
\newblock


\bibitem[Zheng et~al\mbox{.}(2025)]%
        {zheng2025deepresearcher}
\bibfield{author}{\bibinfo{person}{Yuxiang Zheng}, \bibinfo{person}{Dayuan Fu}, \bibinfo{person}{Xiangkun Hu}, \bibinfo{person}{Xiaojie Cai}, \bibinfo{person}{Lyumanshan Ye}, \bibinfo{person}{Pengrui Lu}, {and} \bibinfo{person}{Pengfei Liu}.} \bibinfo{year}{2025}\natexlab{}.
\newblock \bibinfo{title}{DeepResearcher: Scaling Deep Research via Reinforcement Learning in Real-world Environments}.
\newblock
\showeprint[arxiv]{2504.03160}


\bibitem[Zhou et~al\mbox{.}(2025)]%
        {zhou2025towards}
\bibfield{author}{\bibinfo{person}{Yuchen Zhou}, \bibinfo{person}{Jiayu Tang}, \bibinfo{person}{Xiaoyan Xiao}, \bibinfo{person}{Yueyao Lin}, \bibinfo{person}{Linkai Liu}, \bibinfo{person}{Zipeng Guo}, \bibinfo{person}{Hao Fei}, \bibinfo{person}{Xiaobo Xia}, {and} \bibinfo{person}{Chao Gou}.} \bibinfo{year}{2025}\natexlab{}.
\newblock \bibinfo{title}{Where, What, Why: Towards Explainable Driver Attention Prediction}.
\newblock
\showeprint[arxiv]{2506.23088}


\bibitem[Zhou et~al\mbox{.}(2023)]%
        {zhou2023multimodal}
\bibfield{author}{\bibinfo{person}{Yangming Zhou}, \bibinfo{person}{Yuzhou Yang}, \bibinfo{person}{Qichao Ying}, \bibinfo{person}{Zhenxing Qian}, {and} \bibinfo{person}{Xinpeng Zhang}.} \bibinfo{year}{2023}\natexlab{}.
\newblock \showarticletitle{Multimodal fake news detection via clip-guided learning}. In \bibinfo{booktitle}{\emph{2023 IEEE international conference on multimedia and expo (ICME)}}. IEEE, \bibinfo{pages}{2825--2830}.
\newblock


\bibitem[Zhu et~al\mbox{.}(2025)]%
        {zhu2025internvl3}
\bibfield{author}{\bibinfo{person}{Jinguo Zhu}, \bibinfo{person}{Weiyun Wang}, \bibinfo{person}{Zhe Chen}, \bibinfo{person}{Zhaoyang Liu}, \bibinfo{person}{Shenglong Ye}, \bibinfo{person}{Lixin Gu}, \bibinfo{person}{Hao Tian}, \bibinfo{person}{Yuchen Duan}, \bibinfo{person}{Weijie Su}, \bibinfo{person}{Jie Shao}, {et~al\mbox{.}}} \bibinfo{year}{2025}\natexlab{}.
\newblock \bibinfo{title}{InternVL3: Exploring Advanced Training and Test-Time Recipes for Open-Source Multimodal Models}.
\newblock
\showeprint[arxiv]{2504.10479}


\end{thebibliography}

\end{document}